\documentclass[sigconf]{aamas}

\usepackage{algorithm}
\usepackage{algorithmic}
\usepackage{balance} 
\usepackage{subfigure}
\usepackage{multirow}
\usepackage{mathrsfs}
\usepackage{amsmath}
\usepackage{pifont}
\usepackage{balance}

\newcommand{\squishlist}[1][$\bullet$]
{
    \begin{list}{#1}
    {
        \setlength{\itemsep}{0pt}
        \setlength{\parsep}{2pt}
        \setlength{\topsep}{2pt}
        \setlength{\partopsep}{0pt}
        \setlength{\leftmargin}{1.5em}
        \setlength{\labelwidth}{1.5em}
        \setlength{\labelsep}{0.5em}
    }
}

\newcommand{\squishend}{\end{list}}

\setcopyright{ifaamas}
\acmConference[AAMAS '24]{Proc.\@ of the 23rd International Conference
on Autonomous Agents and Multiagent Systems (AAMAS 2024)}{May 6 -- 10, 2024}
{Auckland, New Zealand}{N.~Alechina, V.~Dignum, M.~Dastani, J.S.~Sichman (eds.)}
\copyrightyear{2024}
\acmYear{2024}
\acmDOI{}
\acmPrice{}
\acmISBN{}

\acmSubmissionID{749}

\title{Mutual Information as Intrinsic Reward of Reinforcement Learning Agents for On-demand Ride Pooling}

\author{Xianjie Zhang}
\affiliation{
  \institution{Dalian University of Technology}
  \city{Dalian}
  \country{China}
  }
\email{xj_zh@foxmail.com}

\author{Jiahao Sun}
\affiliation{
  \institution{Dalian University of Technology}
  \city{Dalian}
  \country{China}
  }
\email{Start1w@mail.dlut.edu.cn}

\author{Chen Gong}
\affiliation{
  \institution{University of Virginia}
  \city{Virginia}
  \country{USA}
  }
\email{chengong@virginia.edu}

\author{Kai Wang}
\affiliation{
  \institution{Nanyang Technological University}
  \country{Singapore}
  }
\email{kai_wang@ntu.edu.sg}

\author{Yifei Cao}
\affiliation{
  \institution{Dalian University of Technology}
  \city{Dalian}
  \country{China}
  }
\email{yfcao@mail.dlut.edu.cn}

\author{Hao Chen}
\affiliation{
  \institution{Institute of Automation, Chinese Academy of Sciencec}
  \city{Beijing}
  \country{China}
  }
\email{hao_universe@163.com}

\author{Yu Liu$^\ast$}\thanks{*Corresponding author}
\affiliation{
  \institution{Dalian University of Technology}
  \city{Dalian}
  \country{China}
  }
\email{yuliu@dlut.edu.cn}

\begin{abstract}
The emergence of on-demand ride pooling services allows each vehicle to serve multiple passengers at a time, thus increasing drivers' income and enabling passengers to travel at lower prices than taxi/car on-demand services (only one passenger can be assigned to a car at a time like UberX and Lyft). 

Although on-demand ride pooling services can bring so many benefits, ride pooling services need a well-defined matching strategy to maximize the benefits for all parties (passengers, drivers, aggregation companies and environment), in which the regional dispatching of vehicles has a significant impact on the matching and revenue. 
Existing algorithms often only consider revenue maximization, which makes it difficult for requests with unusual distribution to get a ride. 
How to increase revenue while ensuring a reasonable assignment of requests brings a challenge to ride pooling service companies (aggregation companies). 
In this paper, we propose a framework for vehicle dispatching for ride pooling tasks, 
which splits the city into discrete dispatching regions and uses the reinforcement learning (RL) algorithm to dispatch vehicles in these regions. 
We also consider the mutual information (MI) between vehicle and order distribution as the intrinsic reward of the RL algorithm to improve the correlation between their distributions, 
thus ensuring the possibility of getting a ride for unusually distributed requests.
In experimental results on a real-world taxi dataset, we demonstrate that our framework can significantly increase revenue up to an average of 3\% over the existing best on-demand ride pooling method. 
\end{abstract}


\newcommand{\BibTeX}{\rm B\kern-.05em{\sc i\kern-.025em b}\kern-.08em\TeX}

\begin{document}


\pagestyle{fancy}
\fancyhead{}


\maketitle 


\section{Introduction}

With the development of the mobile internet and sharing economy, it is becoming more and more accepted for people to hail a ride anytime by using mobile devices. On-demand ride pooling is one of the most popular services among them, where service providers like UberPool, LyftLine, and GrabShare allow multiple passengers traveling in the same direction to be matched with the same vehicle through intelligent algorithms and smart terminal devices~\cite{rideshare-stats,uberpool-stats,gong2022mind,zhang2023}. The emergence of ride pooling services not only reduces energy consumption and emissions, but also reduces traffic congestion, and lowers the cost of individual passenger's taxi fares. At the same time, it also brings economic benefits to drivers and aggregation companies~\cite{shah2020neural,xianjie2023AAAI}.

However, there are still some challenges in city-scale on-demand ride pooling systems. First, travel demand is not uniformly distributed over different regions of the city, and ride pooling systems often face problems such as an imbalance between supply and demand in different regions. This makes it necessary for the on-demand ride pooling system to reasonably dispatch vehicles to allow as many vehicles as possible to serve the unusually distributed passengers and to reduce the pick-up distances and times of the vehicles. 
Secondly, the ride pooling system matches vehicles and requests according to the position of the vehicles after dispatching. The result of the system dispatching determines the range of order areas that vehicles can choose during the matching process, and the expected revenue of matching is also fed back to the dispatching decision of the system, and there is a dependency relationship between dispatching and matching.
Finally, unlike usual ride sharing, an on-demand ride pooling system requires combining passengers on the same route into a ``trip'' (a combination of requests) and matching them to the same vehicle~\cite{Alonso-Mora462}, which turns the bipartite matching problem of passengers and vehicles into a tripartite graph problem of requests, trips, and vehicles, which is a difficult problem.

Since vehicle dispatching and matching can interact with each other, much work has emerged recently on the optimization of dispatching and matching to improve the overall efficiency of on-demand rides. Some traditional algorithms achieve minimum waiting time and cruising time by routing planning and the nearest matching of vehicles and requests~\cite{RopkeC09,surveyRitzinger}. Some algorithms use combinatorial optimization to improve the success rate of orders~\cite{Zhang17,ma2013t,Lowalekar2019ZACAZ,Alonso-Mora462}. However, these algorithms are computationally expensive, myopic, and ignore potential future impacts. Reinforcement learning has recently been used to solve related problems, considering potential future implications such as~\cite{MARLODM} and CoRide~\cite{CoRide}. Moreover, these algorithms can only optimize the situation where a single vehicle serves a single passenger, and can not be extended to solve the problem of on-demand ride-pooling. In solving the case of ride pooling, NeurADP~\cite{shah2020neural} uses an approximate dynamic programming (ADP) algorithm to consider the future impact of matching, but does not consider the spatial distribution of vehicles and passengers.

To make the algorithm applicable to on-demand ride pooling, and to consider future benefits and the vehicle-order distribution differences while dispatching and matching decisions, we propose a reinforcement learning-based (RL-based) vehicle dispatch framework. The algorithm uses mutual information (MI) between vehicle distribution and request distribution as the intrinsic reward value for vehicle dispatch. Specifically, our contributions are as follows:

\squishlist
    \item We provide a precise definition of the dispatching and matching problem for ride pooling systems. Based on this definition, for the dispatching problem, the city map is divided into an appropriate number of hexagonal regions to facilitate near-field dispatching in these hexagonal regions.
    \item We propose a reinforcement learning-based regional dispatching algorithm that uses a mean field Q-learning (MFQL)~\cite{pmlr-v80-yang18d}, allowing the algorithm to scale up to city-scale ride pooling tasks. 
    \item Optimizing the MI between the distribution of vehicles and requests can enable the ride pooling system to adjust the distribution of vehicles according to the distribution of requests, thereby improving the overall revenue. By using MI as the intrinsic reward value in reinforcement learning (RL), we can optimize the value of MI.
    \item In experiments, we use a simulation of an on-demand pooling task with a real-world dataset to show that our framework represents a 3\% relative improvement over the best available on-demand ride pooling approach.
\squishend

\begin{table}
{\small \begin{tabular}{|c|l|}
    \hline
     Symbol & Definition  \\
     \hline
     \hline
     $V$ & Available vehicles in the current time period\\
     \hline
     $E$ & Batched requests in the current time period\\
     \hline
     $e$ & Variable to denote a request \\
     \hline
     $r$ & Reward value for agent (vehicle) \\
     \hline
     $v$ & Variable to denote a vehicle\\
     \hline
     $f$ & Variable to denote a combination of requests.  \\
     & $E^f$ corresponds to requests in $f$\\
     \hline
     $x^f_v$ & Binary decision variable that is set to 1 if\\ 
     & $v$ is assigned to $f$\\
     \hline
     $\mu$ & Pricing vector for all requests at a time step \\
     \hline
     $o_v^{f}(\mu^f)$ &   
     Expected reward obtained by assigning request \\
     &  combination ${f}$ to $v$ given price vector $\mu^f$\\
     \hline
     $C(.)$ & Matching constraints required for \\
     & feasible matching \\
     \hline
     $a_v$ & Action (dispatching region) for vehicle $v$ \\
     \hline
    $\tau$ &  Pick up delay constraint \\
    \hline
    $\lambda$ & Detour delay constraint \\
    \hline
    $\omega_v$ & Local observation of agent $v$ \\
    \hline
\end{tabular}}
\caption{Summary of the major notations in this paper}
\vspace{-5mm}
\end{table}

\section{Ride-pool Dispatching and Matching Problem (RDMP)}
We can divide this problem into two major parts, one is vehicle dispatching and the other is vehicle-request matching. In the following, we will introduce the overall process at first, and then describe the two main parts in detail separately.

In this problem, we have a set of vehicles $V$. Each vehicle in the set may serve multiple passages (max $c$) at the same time (e.g., UberPool and GrabShare). Passengers' requests are sent to the aggregation platform, and usually, we process a batch of requests simultaneously at a fixed time interval $\Delta$, and the processed requests are called the set $R$. The aggregation platform first dispatches the vehicles to the appropriate regions. Subsequently, the dispatched vehicles are conditionally filtered and need to meet some conditions to form a vehicle set $V$. The conditions can be the vehicle pickup time ($\tau$) of passengers and the detour delay ($\gamma$) of the vehicle to transport passengers. Finally, the aggregation platform matches vehicles with passengers in the optional range and assigns a number of passengers to each vehicle that does not exceed the maximum capacity ($c$). Our goal is to design a dispatching strategy that ensures maximizing the overall expected revenue during the matching phase.

\subsection{Matching Problem}

Without considering the dispatching of vehicles, the on-demand ride pooling can be viewed as a tripartite graph problem of matching among vehicles, trips, and requests.

\noindent \textbf{Matching in a Tripartite Graph:}
The goal of matching is to create a mapping between requests and vehicles and to maximize the overall revenue under constraints. There are two types of taxi services matching cases: (1) Taxi/car on-demand services (ride-hailing), in which the problem can be viewed as ``maximum weight bipartite matching"; (2) On-demand ride-pooling services, in which bipartite matching can not capture the structure of the matching problem and a vehicle receives multiple requests at the same time.
For the ride pooling, a vehicle receives both A and B requests, but these two requests may result in unacceptable detour delays, and this type of combination constraint cannot be captured in the bipartite graph, making it impossible for the vehicle to serve both requests.
We can solve this problem by creating an intermediate representation called a ``trip", which is a combination of requests denoted by $f$. Then a mapping is formed between vehicles and requests by trips, and the associated weights of the mapping are $ o_v^f $. This mapping relationship ensures that vehicles and requests are assigned at most one trip, and the maximum weight matching in the tripartite graph can be seen as the solution to the matching problem. Use the following equation to construct this optimization:
\begin{equation}
    X^* = \arg\max_{X \in C(E)} \sum_{x^f_v \in X} x^f_v \cdot o^f_v \label{eqn:2}
\end{equation}
where, $x_v^f$ is a binary decision variable indicating whether vehicle $v$ chooses trip (request combination) $f$ or not. $o_v^f$ indicates the revenue of vehicle $v$ in the case of choosing trip $f$. $C(E)$ is the constraint of feasible matching.

We utilize the objective function to perform matching operations by solving an integer linear programming to (ILP). \citet{Alonso-Mora462} provides an approximation method that solves the problem of exponentially increasing the number of trips, and this method generates feasible trips that work well in practice and satisfy the constraint of the requests (described in the next section).

The objective function in the formula is  $o_v^f$, which can be specifically profit/revenue, a fairness indicator~\cite{lesmana2019balancing}, or a way to integrate future information~\cite{shah2020neural}.
Each of these objectives is likely to be a system-level goal that the ride pooling platform wants to maximize, and all of them can be modeled as linear functions of trip revenue ($\alpha$ and $\beta$ are constants):
\begin{equation} \label{eqn:linearmatch_}
    o^f_v = \alpha^f_v \cdot {\sum}_{e \in E^f} \mu^e + \beta^f_v
\end{equation}
The overall objectives depend on the existing trajectory of the vehicle $v$ and the source and destination of all requests in the trip $f$. $\mu$ is price for request.

\subsection{Dispatching  Problem}

The current state of the dispatched vehicle is related to the previous moment state so that the dispatch can be viewed as a sequential decision task and RL can be used to obtain a dispatch policy for the vehicles. Using a centralized dispatching approach is not realistic, since each city has thousands of vehicles running at the same time. Using a multi-agent approach to decompose the centralized actions into local actions of individual vehicle is one option~\cite{zhang2021,zhang2022}. In addition, each vehicle is only concerned with local request information in the surrounding area, and the problem can be modeled as Partially Observable Markov Decision Process (POMDP)~\cite{spaan2012partially}.

The  dispatching  framework  considers  POMDP  as  a  tuple, which can be written as $\langle S,\Omega,Z,P,A,R,D,N,\dot{\gamma} \rangle$. In the tuple, $S$ represents the global state, $P$ represents the state transfer function, $A$ represents the action set, $R$ represents the reward function, $D$ represents the grid set, $N$ represents the number of agents, and $\dot{\gamma}$ represents the discount factor for future rewards. For each agent (vehicle) $v$, $\omega_v\in \Omega$ denotes the local observation of the vehicle, and the local observation set $\Omega$ is obtained through the observation function $Z(S,v)$. 
The dispatching action for each vehicle $v$ is $a_v \in A$ and the reward value is $r_v\in R$. $D_v\in D$ denotes the grid region where the vehicle $v$ is located. Once the decision is made, the state transition occurs, i.e., the agents execute their actions, and the state $S_t$ of the environment at time $t$ changes to $S_{t+1}$, and the vehicle $v$ receives its reward $r_v^t$. The goal of each agent in the above definition is to maximize the cumulative reward $G_{t:T}$ from $t$ to $T$.


$$
\max G_{t: T}=\max \sum_{t=0}^T \dot{\gamma}^t r_v^t
$$
where $\dot{\gamma}$ is the discount factor.

\begin{figure}[ht]
    \centering
    \includegraphics[width=1.6in,height=2.1in]{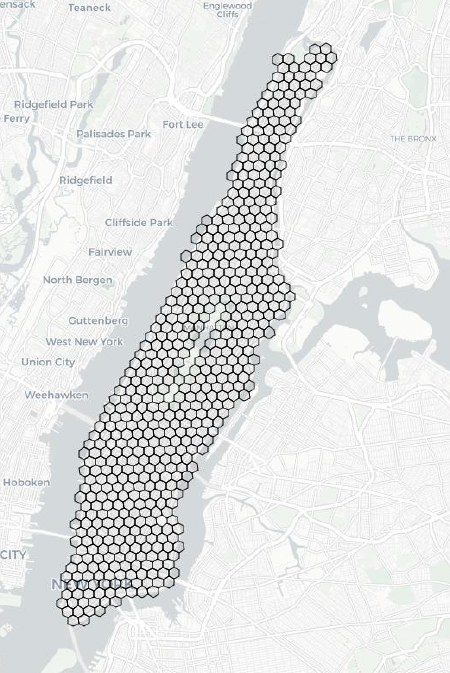}
    \caption{The Manhattan city area is divided into small dispatching regions by hexagonal grids. The grid is divided using the tool h3 (https://h3geo.org) with a resolution of 8.}
    \label{fig:Regions}
\end{figure}

The city is divided into discrete regions using hexagonal grids, each grid is a vehicle dispatch region, and each grid contains multiple requests and vehicles. As shown in Figure~\ref{fig:Regions}, it is the result of grids that divide the urban area of Manhattan. Based on the above multi-agent RL settings, we mathematically formulate the definition of the vehicle dispatching task as follows:
\squishlist
\item[\textbf{State:}] for each vehicle $v$ the local observation can be represented as a tuple of six elements, i.e. 
$\omega_v = \langle l_v, l_{M_v}, e_v, E_{M_v}, V_{M_v}, t \rangle$. The elements in the tuple are $l_v$ for the vehicle location, $l_{M_v}$ for the vehicle location information in the neighboring grids (${M_v}$ represents the grid within the range considered by the vehicle), $e_v$ for the existing passengers on vehicle $v$, $E_{M_v}$ for the number of orders in the neighboring grids, $V_{M_v}$ for the number of vehicles in the neighboring grids, and $t$ for the time. At time $t$, the values of $E_{M_v}$, $V_{M_v}$ are the same for all vehicles in the same grid except for the vehicle's own characteristics $l_v$, $l_{M_v}$, and $e_v$.

\begin{figure}[ht]
    \centering
    \includegraphics[width=1.4in,height=1.4in]{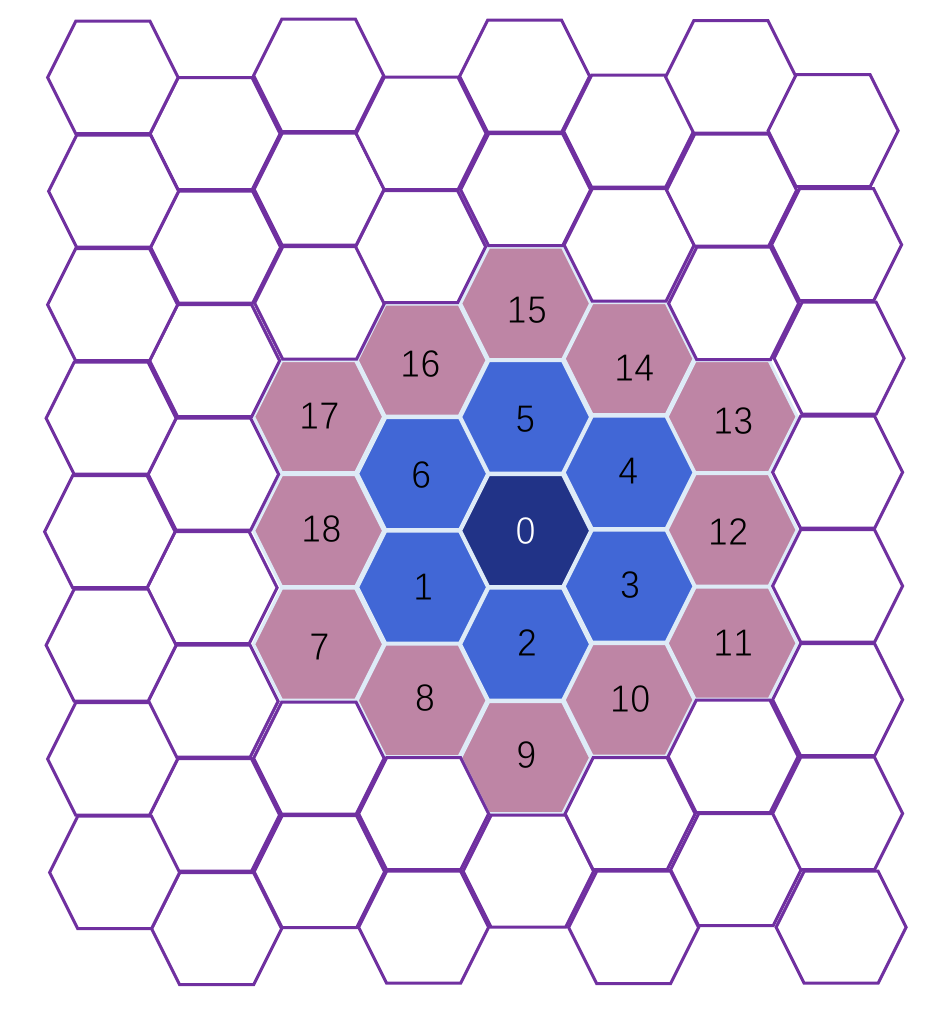}
    \caption{The figure shows the action setting of the agent and also the dispatching region set for the vehicle. Vehicle in region $d_0$ have candidate dispatching regions as $\{d_0, \dots , d_{18}\}$.
    }
    \label{fig:Actions}
\end{figure}

\item[\textbf{Action:}] We denote the action set of vehicle $v$ as $A=\{d_i\}_{i\in M_v }$. $\{d_i\}_{i\in M_v }$ are the near regions in which vehicle $v$ can be nearly dispatched, and $a_v \in A$ denotes the dispatch region selected by vehicle $v$. For example, Figure~\ref{fig:Actions} shows that the vehicle location is $d_0$, then its candidate dispatching action set is the regions $\{d_0,\dots,d_{18}\}$. Because a far dispatch will increase the pickup time of the vehicle, the area of two layers around the vehicle location is set as the optional dispatch region, as shown in Figure~\ref{fig:Actions}. The pickup distance within this range is reasonable (The diameter of the hexagon is 0.36 km), and the two adjacent layers allow more optional dispatch regions for vehicles. All the vehicles in our setup are homogeneous, and the surrounding arrangement of actions of other agents is consistent.

\item[\textbf{Reward:}] The vehicle $v$ is dispatched to the region pointed by the action and matched with the combination of requests in the region. The reward value is finally obtained based on the revenue value of the requests after matching. A reward function proportional to the price of requests is designed. In addition, to satisfy the system's consideration of unusually distributed requests, the MI between vehicle distribution and request distribution is added as an intrinsic reward value (detailed in the next section).
\squishend

\section{RL-based dispatching and matching framework}\label{joint-matching-and-pricing}

The simple matching algorithm can only get the vehicle-request matching policy with the highest revenue at the current moment, which is a rather myopic algorithm. To avoid this myopia, in our framework, vehicles are modeled as RL agents to obtain the long-term future impact of the vehicle dispatching strategy. The dispatched vehicles are then involved in vehicle-request matching, so that both the long-term view of the dispatching strategy and the maximum revenue of timely matching are taken into account in our algorithm.

Specifically, the framework includes the following components:
\squishlist
\item The Q-learning is used to obtain the vehicle dispatching strategy. The Q function calculates the Q-value $Q(\omega_v,a_v)$ of all possible actions of vehicle $v$. The Q-value is used to select the action of $v$.

\item Once the action $a_v$ is determined, the vehicle $v$ will dispatch to the region $d_v$, and the vehicle $v$ is matched with the requests in the region $d_v$.

\item To further capture the dependence of dispatching actions on neighboring vehicles/agents, we use MFQL in the RL dispatching framework.

\item To better serve the unusually distributed requests, the mutual information between the request and vehicle distribution is added as a reinforcement learning intrinsic reward value.
\squishend
The overall process of the framework is shown in Appendix Algorithm~\ref{alg:algorithm1}.

\subsection{MFQL}
\label{MFQL}
In MFQL, the interactions between agent populations can be viewed as interactions between the individual agent and the average influence of neighboring agents. MFQL has been successfully applied to a large number of agent problems with aggregated actions, and similarly, we adopt this idea in computing neighborhood vehicle dispatching regions.

The state, actions, and reward values are defined using the problem in the previous section.
The action (dispatching region) ${a}_{v}$ for each vehicle $v$ is a discrete categorical variable, denoted as a one-hot encoding. The state provides the distribution of vehicles and requests in the current agent's surrounding region. The actions of neighboring agents provide a signal of supply demand imbalance in the surrounding regions of the neighbor. Specifically, the average action-average response of $N_v$ of the neighboring agents of vehicle $v$ is ${\overline{a}}_{N_v} = \frac{1}{|N_{v}|}\sum_{k \in N_v}^{}\mspace{2mu} a_{k}$. According to the assumptions of the MFQL, vehicle $v$ is only affected by the actions of neighboring vehicles, so the Q function can be written as $Q_{v}\left( \omega_{v},{\overline{a}}_{N_{v}},a_{v} \right)$.

In the selection of vehicle dispatching actions, the Boltzmann softmax selector is used to obtain the final dispatching action probabilities.
\begin{equation}
\begin{split}
    a_v & \sim \pi_{v}( \cdot \mid \omega_{v},{\bar{a}}_{N_v}), \\
    \pi_{v}( \! a_{v} \! \mid \! \omega_{v},{\bar{a}}_{N_v}\!) & \!=\! \frac{\exp{\left(  Q_{v}\left( \omega_{v}, a_{v},{\bar{a}}_{N_v} \right)/ \Gamma \right)}}{\sum_{a_{v}' \in \mathcal{A}_v} \! \exp \! \left( \beta \cdot Q_{v}\left( \omega_{v},a_{v}',{\bar{a}}_{N_{v}} \right)/ \Gamma \right)} 
\end{split}
\label{eqn:actionsel}
\end{equation} 
where $\Gamma$ is a non-negative temperature parameter, which is not explored at all when $\Gamma=0$ and randomly selected for actions when $\Gamma \to \infty$.
The mean field Q function is updated as follows:
\begin{equation}
    Q^{t+1}_{v}\!\left(\!\omega, a_{v}, \bar{a}_{v}\!\right)\!\!=\!(\!1-\dot{\alpha}\!) Q^{t}_{v}\left(\omega, a_{v}, \bar{a}_{v}\right)+\dot{\alpha}\!\left[r_{v}\!\!+\!\!\dot{\gamma} {V_{\text{MF}}}^{t}_{v}\left(\omega^{\prime}\right)\!\right]
\end{equation}
\begin{equation}
    {V_{\text{MF}}}^{t}_{v}\left(\omega^{\prime}\right) = 
    \sum_{a_{v}} \pi^{t}_{v} \mathbb{E}_{\bar{a}_{v}\left(\boldsymbol{a}_{-v}\right) \sim \pi^{t}_{-v}}\left[Q^{t}_{v}\left(\omega^{\prime}, a_{v}, \bar{a}_{v}\right)\right]
\label{eqn:mfqlup}
\end{equation}
The $-v$ is the other agent except agent $v$.

\subsection{Intrinsic Reward}
\label{IntrinsicReward}
Considering the request distribution is crucial to the dispatch of vehicles, if the correlation between vehicle distribution and order distribution is increased through the system's dispatch, intuitively it can match the vehicle and request relatively bette.

Specifically, there are two ways to make increase the correlation. The first way is to maximize the entropy value $H[V_D]$ of the vehicle distribution, the more uniformly the vehicles are distributed in the urban regions the higher this entropy value is. The second way is to minimize the conditional entropy $H[V_D |E_D]$ for a given request distribution to make the distribution of vehicles relatively deterministic. According to the mathematical relationship between MI and entropy, $H[V_D]-H[V_D |E_D]=I(V_D;E_D )$, i.e., the above two ways can be achieved by maximizing the MI between the vehicle and the request distribution.

However, estimating and maximizing MI is usually intractable. Taking inspiration from the literature on variational inference~\cite{Martin2008,Alemi2017,gong2020stable}, the variational posterior estimator is introduced to derive a tractable lower bound on the MI for each time step $t$.

Define $p(v_d)$ as the proportion of vehicles in region $d$, and $p(e_d)$ as the proportion of requests in region $d$. Computing $I(V_D;E_D )$ with the distribution of vehicles and orders in each grid as follows:
\begin{equation}
\begin{aligned}
I\left(V_D ; E_D\right) & =\iint p\left(v_d, e_d\right) \log \frac{p\left(v_d, e_d\right)}{p\left(v_d\right) p\left(e_d\right)} \\
& =\iint p\left(v_d, e_d\right) \log \frac{q\left(v_d \mid e_d\right)}{p\left(v_d\right)} \\
& +D_{k l}\left(p\left(v_d \mid e_d\right) \| q\left(v_d \mid e_d\right)\right) \\
& \geq \iint p\left(v_d, e_d\right) \log \frac{q\left(v_d \mid e_d\right)}{p\left(v_d\right)} \\
& =\!\! -\mathbb{E}_{e_d \sim \! E_D} \! \left[CE\!\left[p\left(v_d \! \mid  \! e_d\right) \! \| q\left(v_d \mid e_d\right)\right] \!\! + \!\! H \! \left(v_d\right)\right.
\end{aligned}
\label{eqa:MI}
\end{equation}
where $q(\cdot)$ is the parameterized variational estimation function. Since $D_{kl} (p(v_d |e_d )||q(v_d |e_d ))$ is a KL divergence, which is non-negative, an inequality is obtained in Equation~\ref{eqa:MI}.
The neural network is used in our algorithm to encode the distribution of requests, where we can describe q as an encoder. $CE(\cdot)$ and $H(\cdot)$ are the computational operators for cross-entropy and entropy, respectively.

\begin{figure}[ht]
    \centering
    \includegraphics[width=3.3in]{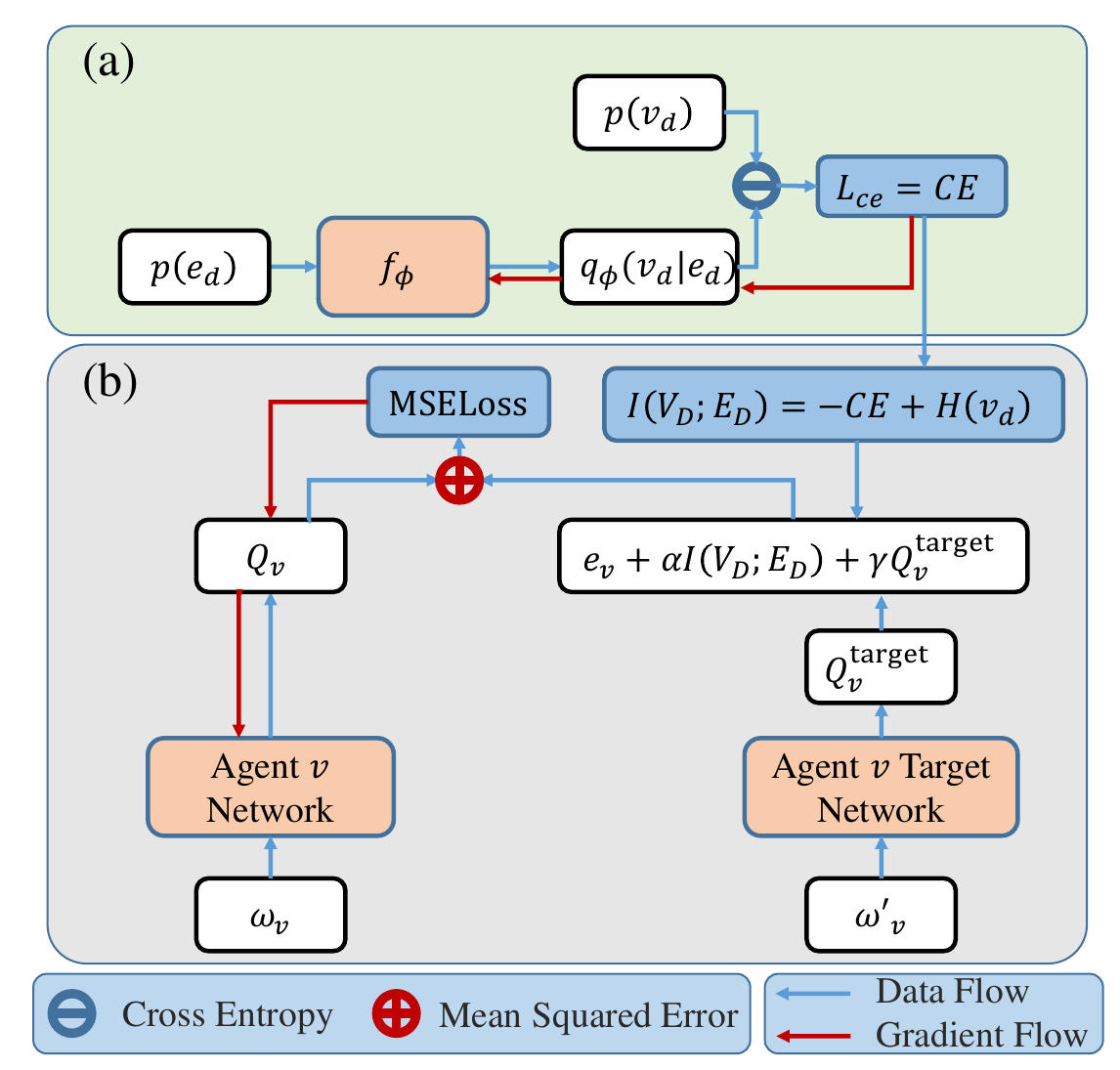} 
    \caption{The overall framework. (a) the process of computing the mutual information of vehicle and request distribution. (b) the training process of Q-learning.}
    \label{fig:Framework}
\end{figure}

As shown in Figure~\ref{fig:Framework}(a), to calculate the variational distribution $q(v_d |e_d )$, $f_\phi$ is set as the variational network in the model. In Figure~\ref{fig:Framework}(b), the cross-entropy $CE \left[ p\left(v_d \! \mid  \! e_d\right) \! \| q\left(v_d \mid e_d\right)\right]$ and the entropy of the vehicle distribution $H (v_d)$ are summed up as an intrinsic reward value for the learning of the agents.
The total reward value function can be rewritten as
\begin{equation}
r_v+\ddot{\alpha} I(V_D;E_D)
\label{eqn:MIalpha}
\end{equation}
where $r_v$ is the reward value obtained from the environmental feedback, and $\ddot{\alpha}$ is the manually set hyperparameter value as a weight to determine the scale of the contribution of the intrinsic reward value $I(V_D;E_D)$ to the total reward value function.

\subsection{Constraint Matching in Dispatched Region}
\label{matching}
The dispatching region $a_v$ of vehicle $v$, obtained using RL, is the policy with forward-looking considerations. The dispatched region of the vehicle can be used as a constraint on the combination of requests matched by the vehicle. Specifically, we add a matching region constraint to the selection of candidate matching requests. Then we can obtain the set $\mathcal{F}_v^t$ of available request combinations for vehicle $v$ as follows.
\begin{equation}
\begin{split}
        \mathcal{F}^{t}_{v} = & \left\{ f_{v} \mid f_{v} \in \cup_{c' = 0}^{c^{v}}{\mathcal{\lbrack U}\rbrack^{c'}},\text{PickUpDelay}( f_{v},v) \leq \tau \right.\  \\
        &\left. \ \text{DetourDelay}( f_{v},v) \leq \lambda, \text{Localtion}(f_{v}) \in a_v \right\}
\end{split}
\label{eqn:reqcomb}
\end{equation}
where, the constraint $\text{Localtion}(f_{v}) \in a_v$ restricts the requests $e_f$ in the available request combination $f_v$ come from the region $a_v$, and each vehicle $v$ is dispatched at time $t$. 
$[\mathcal{U}_v ]^{c'}$ denotes the combination of requests at capacity $c'$, which is a number less than the maximum capacity $c^v$ of the vehicle. $F_v^t$ contains an empty matching combination $f_v=\phi$ when $c'=0$. $\tau$ is the time threshold for picking up the combination of requests and $\lambda$ is the detour delay.

The optimization objective of matching is:
\begin{equation}
   o(x^t) = {\sum_{v}\sum_{f \in \mathcal{F}^{t}_{v}}}o^{t,f}_{v} \cdot x^{t}_{v,f}
   \label{eqn:ILPobj}
\end{equation}
The matching also requires additional constraints to be satisfied. 
\squishlist
\item Each vehicle $v$ is assigned only one request combination (containing empty request combinations) (Equation~\ref{cons:a1}).
\item Each request can only be assigned to no more than one vehicle (Equation~\ref{cons:b2}). 
\item Vehicles can be assigned or not assigned request combinations $f$ (Equation~\ref{cons:c3}).
\squishend
\begin{align}
& \sum_{f \in {\mathcal F}_{v}^{t}} x_{v}^{t,f} = 1 ::: \forall v \label{cons:a1} \\
& \sum_{v} \sum_{f \in {\mathcal F}_{v}^{t};r \in f} x_{v}^{t,f} \leq  1 ::: \forall r \label{cons:b2} \\
& x_{v}^{t,f} \in \{0,1\} ::: \forall v,f  \label{cons:c3}
\end{align}

\section{Experimental Results}

This section introduces the experimental setup and the performance results of the overall methods for the benchmark simulations based on a real dataset at the city scale.


\subsection{Setup}

\subsubsection{Dataset description}
To make the simulation environment close to the real-world taxi scenario, we chose the street network of Manhattan, New York as the source of dependency for vehicle operation. The New York yellow taxi dataset~\cite{yellowtaxi} is also obtained from the open network. We use ~\citet{boeing2017osmnx}'s work osmnx to obtain the city's street network from openstreetmap using ``drivers". 
Our experimental setup is similar to papers~\cite{xianjie2023AAAI,shah2020neural} in terms of street network setup where the street network is a graph structure $\mathcal{G}$. In $\mathcal{G}$, the vertices are the street intersections and the edges are the streets connecting these intersections.
The original road is processed, and the network nodes that have no outgoing degree are deleted. Finally, a street network with 4373 vertices and 9540 edges is generated. Since only Manhattan urban area are considered, the filtered orders with both pick-up and drop-off locations in Manhattan account for roughly 75\% of the total. Our final dataset contains an average of 322714 requests per weekday and 19820 requests during peak hours. 

\begin{table*}[htbp]
\centering
\setlength\tabcolsep{3pt}
\begin{tabular}{lllllllllll} 
    \hline
    \multirow{2}{*}{Varying} & \multicolumn{3}{c}{Parameters} & \ding{172}  & \ding{173}  & \ding{174}   & \ding{175}  &\multirow{2}{*}{\ding{174}/\ding{173}}	&\multirow{2}{*}{\ding{175}/\ding{173}}	&\multirow{2}{*}{\ding{175}/\ding{174}} \\
    \cline{2-4}
                                     & NV    & PD  & C   &Random &NOD &DQN &DQN+MI\\
    \hline
    \multirow{3}{*}{PD}       &  1000	   &  100	 & 4	& 92298.79 	      &   116624.39	 &  122797.35 &  124775.29  & 5.29\% & 6.99\%	&1.61\%\\
                                        &  1000	   &  200	 & 4	& 119873.25 	  &   134910.56	 &  158340.14 &  161070.32  &17.37\%	&19.39\%	&1.72\%\\
                                        &  1000	   &  300	 & 4	& 131347.95 	  &   143365.55  &  170624.32 &  174132.36  &19.01\%	&21.46\%&	2.06\%\\
    \hline                                                                                                                 
    \multirow{3}{*}{NV} &  600	   &  300	 & 4	& 92852.57 	      &   98001.92	&  107571.97 &  110672.99  &9.77\%	&12.93\%	&2.88\%\\
                                        &  800	   &  300	 & 4	& 108449.56 	  &   121868.77	&  141061.41 &  144028.22  &15.75\%	&18.18\%	&2.10\%\\
                                        &  1000	   &  300	 & 4	& 131347.95 	  &   143365.55	&  170624.32 &  174132.36  &19.01\%	&21.46\%	&2.06\%\\
    \hline                                                                                                                 
    \multirow{3}{*}{C}           &  1000	   &  300	 & 2	& 107210.89 	  &   118332.23	&  124867.07 &  126988.39  &5.52\%	&7.32\%	&1.70\%\\
                                        &  1000	   &  300	 & 4	& 131347.95 	  &   143365.55	&  170624.32 &  174132.36 & 19.01\%	&21.46\%	&2.06\%\\
                                        &  1000	   &  300	 & 8	& 140376.62 	  &   146901.66	&  189352.47 &  194625.15  &28.90\%	&32.49\%	&2.78\%\\
    \hline
    
    \hline
\end{tabular}
\caption{The total revenue in peak hours of 18-19, the effect of different dispatching and matching strategies on the results. Here the abbreviations ``NV'', ``PD'', and ``C'' are ``number of verhicle'', ``pickup delay'', and ``capacity'', respectively. \ding{175}/\ding{174} is the percentage increase of the algorithm \ding{175} with respect to \ding{174}.}
\label{tab:pricing_impact_DQN}
\end{table*}

The dataset, for each request, is well described in the following specific attributes: pick-up and drop-off locations (latitude and longitude), and request pick-up and drop-off times (specific to different times of the day, and times of the week). Based on this dataset and the modeled street network, we make the following main settings:
\squishlist
\item Requests are mapped to the nearest available street intersections in the network, based on the latitude and longitude coordinates where they are located.

\item The request pickup times, based on each batch of requests interval $\Delta$, are transformed into decision times.

\item The travel time of the street uses the average daily travel time in the method~\cite{santi2014quantifying} as an estimate of the travel time.

\item To facilitate the dispatching of the nearest region of vehicles, We take the Manhattan map and divide it into hexagonal grids with a resolution of 8 according to the resolution table \footnote{https://h3geo.org/docs/core-library/restable}, as shown in Figure~\ref{fig:Regions}. These grids are divided by the tool h3\footnote{https://h3geo.org}.
\squishend

\subsubsection{Simulation Engine}
The request data is static dataset, while the RL agents interact with a dynamically changing environment. Our simulation engine is to build a dynamically updated simulation environment that can interact with the agents (vehicles) based on these static request data.

First, the simulation engine generates temporal requests based on the time information and starting location of the original request. Second, we initialize randomly located vehicles to pick up the requests matched by the algorithm and deliver the requests to the request's drop-off location. We initialize the locations of all vehicles only once at the beginning (a fixed number of vehicles with a fixed carrying capacity). Subsequent updates of the vehicle positions are based on the simulator for the vehicle carrying process and the cruise process.

The dispatching and location update of vehicles is controlled by both the RL algorithm and the simulator.
In the dispatching phase, the RL algorithm gives the dispatching policy for the vehicles. In the matching phase, the simulator puts the requests into the corresponding vehicles according to the obtained matching policy.
The simulator then moves the vehicles through the street network based on the shortest travel time between the requests' pickup and drop-off locations, and the location updates of all vehicles are strictly based on the accumulated travel time to determine the current location of the vehicles.
Our simulator can be divided into the following main modules:
\squishlist
\item[\textbf{Vehicle status update module:}] Update the current vehicle position at each time step according to the vehicle travel time.

\item[\textbf{Request assignment module:}] According to the matching result obtained by the algorithm, the corresponding request is assigned to the specified vehicle and the vehicle status is updated.

\item[\textbf{Regional dispatch module:}] Based on the dispatch actions obtained by the RL dispatch algorithm, the system matches requests with vehicles based on the dispatch region.

\item[\textbf{Vehicle routing module:}] The simulator integrates a vehicle closest route generation module, so that once the destination is known the vehicle will travel along the shortest path.
\squishend

\begin{table*}[htbp]
\centering
\setlength\tabcolsep{1pt}
\begin{tabular}{lllllllllllll} 
    \hline
    Coefficient of MI ($\ddot{\alpha}$) & 0.005	&	0.01	&	0.03	&	0.05	&	0.07	&	0.09	\\
    Total revenue &  	173742.01	&	173533.79	&	175778.08		&176700.82	&	177071.15	&	174625.08 \\
    Request & 9879.4 &  9888.6 & 10001.4 & 10089.6 & 10100.4 & 9960.0\\
    \hline
    Coefficient of MI ($\ddot{\alpha}$) &	0.1	&	0.3	&	0.5	&	0.7	&	0.9	&	1 \\
    Total revenue &	170624.32 	&	164443.39	&	156424.33	&	157549.18	&	146886.46	&	154567.15\\
    Request & 9689.4 & 9358.6 & 8875.4 & 8928.8 & 8327.4 & 8748.8 \\

    \hline
\end{tabular}
\caption{Results under different coefficients ($\ddot{\alpha}$) of MI in Equation~\ref{eqn:MIalpha}. The number vehicles is 1000 and the capacity is 4.}
\label{tab:differentcoeffi}
\end{table*}



\begin{table*}[htbp]
\centering
\setlength\tabcolsep{3pt}
\setlength\tabcolsep{3pt}
\begin{tabular}{lllllllllllll} 
    \hline
    Numbers & {3} & {4} & {5}  & {6}  &7	&8	&9\\
    Revenue		& 173393.79		&174165.75	&173516.57		&174818.37	&172673.65	&171552.36	&173231.41 \\
    Revenue-172000		& 1393.79		&2165.75	 &1516.57		&2818.37	&673.65	&-447.64	&1231.41 \\
    \hline
\end{tabular}
\caption{ Results under different neighbor vehicle numbers. The number vehicles is 1000 and the capacity is 4. }

\label{tab:differentNumberNeighbor}
\end{table*}

\begin{table}[htbp]
\centering
\setlength\tabcolsep{3pt}
\begin{tabular}{lllllllllll} 
    \hline
    \multirow{2}{*}{Varying} & \multicolumn{3}{c}{Parameters} & \multirow{2}{*}{MFQL}  & \multirow{2}{*}{MFQL+MI} &	\multirow{2}{*}{Improve}\\
    \cline{2-4}
                                     & NV    & PD  & C   \\
    \hline
    \multirow{3}{*}{PD}                 &  600	   &  100	 & 4	& 76163.63 	   &  77849.47  	&2.21\% \\
                                        &  600	   &  200	 & 4	& 96342.77 	   &  97959.34  	&1.68\% \\
                                        &  600	   &  300	 & 4	& 108190.70 	   &  110196.19  	&1.85\% \\
    \hline                                                                                                                 
    \multirow{3}{*}{NV}                 &  300	   &  300	 & 4	& 52936.71 	   &  53879.17  	&1.78\% \\
                                        &  400	   &  300	 & 4	& 73321.78	   &  74147.34  	&1.13\% \\
                                        &  500	   &  300	 & 4	& 89143.85 	   &  92292.44  	&3.53\% \\
    \hline                                                                                                                 
    \multirow{3}{*}{C}                  &  600	   &  300	 & 2	& 76370.79 	   &  76693.32  	&0.42\% \\
                                        &  600	   &  300	 & 4	& 108190.70 	   &  111326.63 	&2.90\% \\
                                        &  600	   &  300	 & 8	& 118254.63 	   &  120660.29  	&2.03\% \\
    \hline
    
    \hline
\end{tabular}
\caption{The total revenue, the effect of different dispatching and matching strategies on the results.}

\label{tab:pricing_impactMFQL}
\end{table}

\begin{figure}[htbp]
\centering    
\subfigure[1000 vehicles, capacity of 4.] 
{
	\begin{minipage}{0.47\linewidth}
	\includegraphics[width=2in,height=1.5in]{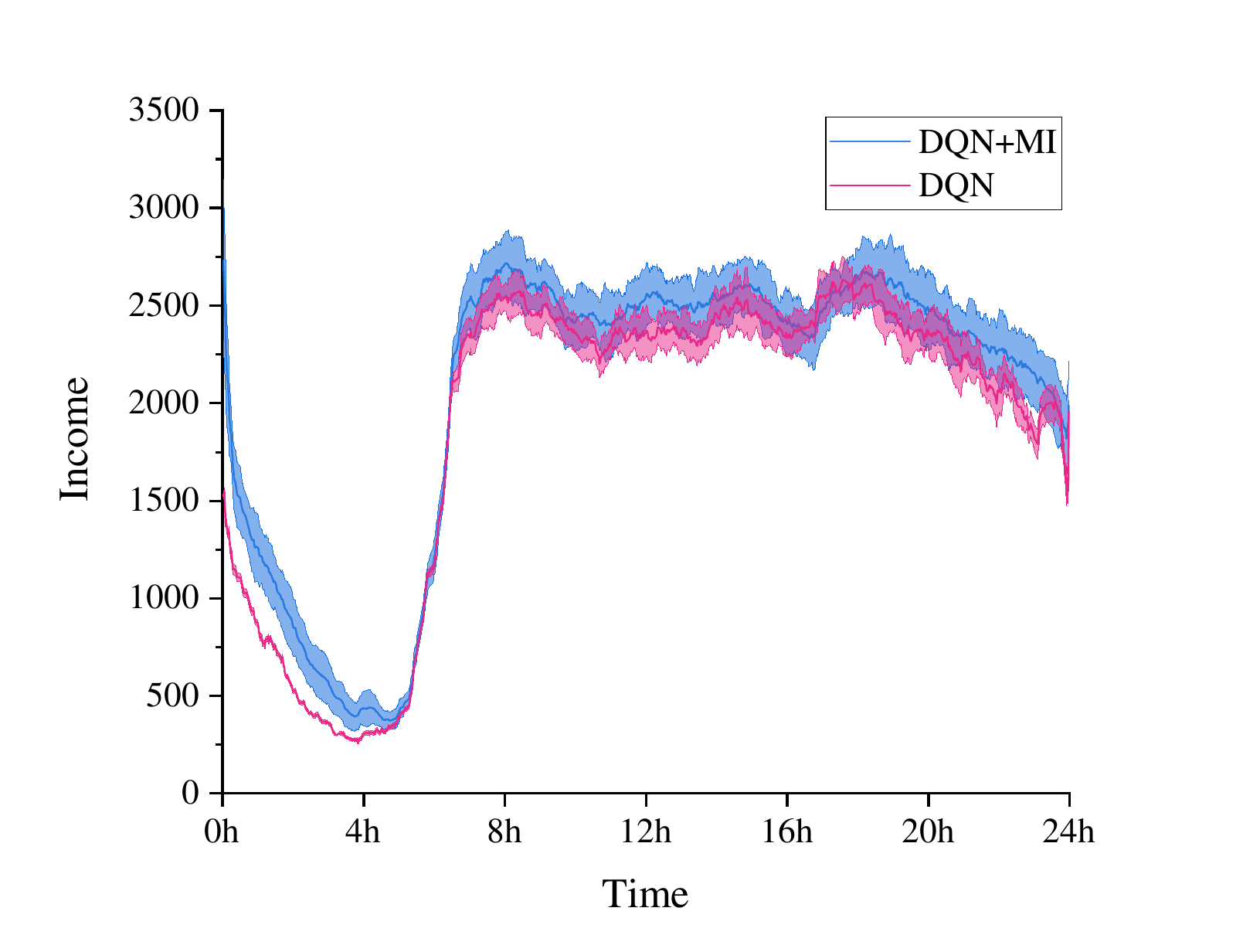}   
	\end{minipage}
}
\subfigure[800 vehicles, capacity of 4.] 
{
	\begin{minipage}{0.47\linewidth}
	\includegraphics[width=2in,height=1.5in]{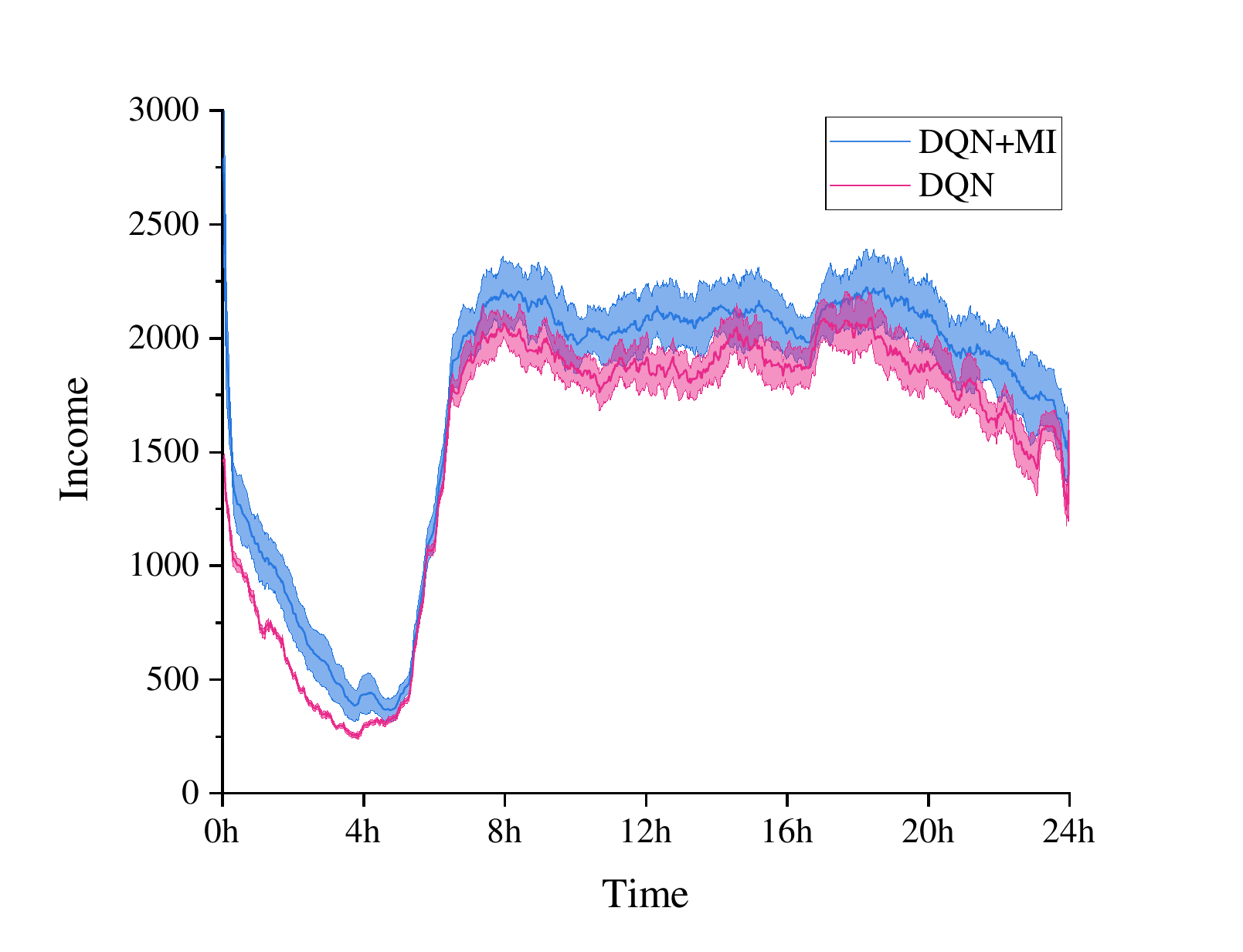}   
	\end{minipage}
}
\caption{The MI module is added to the DQN and the revenue curve of the results of the whole day running.} 
\label{fig:DQN_oneday}  
\end{figure}


\begin{figure}[ht]
    \centering
    \includegraphics[width=3.3in]{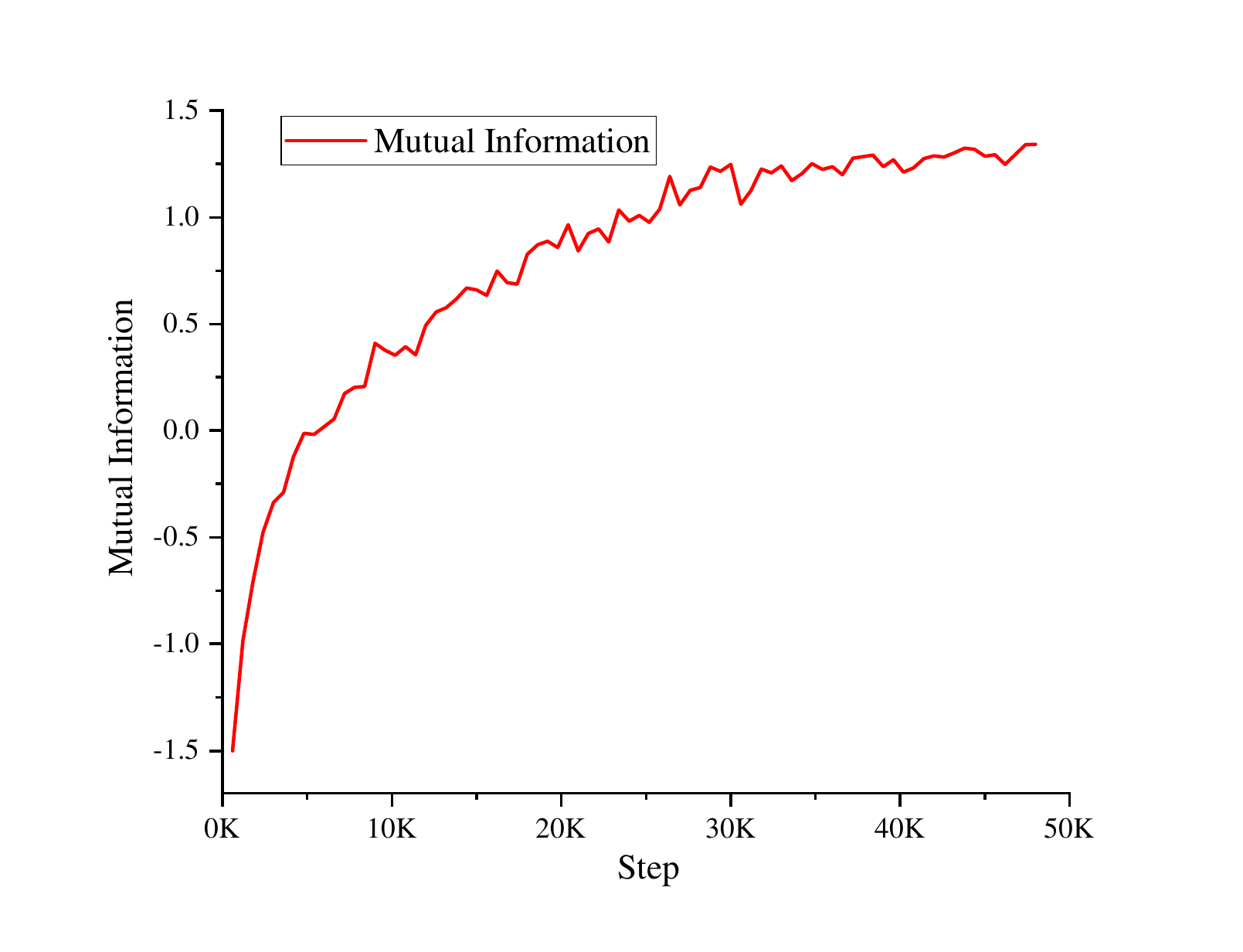}
    \vspace{-5mm}
    \caption{The curve is the mutual information value. We set the number of vehicles at 1000 and capacity at 4.}
    \label{fig:MutualInformation_value}
\end{figure}

\subsection{Baseline}
To demonstrate the effectiveness of our proposed algorithm, the impact of the dispatching policy and the MI of the request and vehicle distribution on the algorithm is verified. We conducted experiments for the case of individual optimization of these components, specifically.

\squishlist
\item To illustrate the effectiveness of our dispatching algorithm based on the Q-learning, the randomized dispatching policy and the nearest request matching policy are compared.

\item To illustrate the impact of the mean field component on the algorithm, DQN and MFQL are compared with the same other components.

\item To illustrate the impact of considering the MI of request and vehicle distribution, the impact of having or not MI on performance is compared, with the same other components.
\squishend

For the above comparison purpose, the following algorithms are set up for experimental testing:
\squishlist
\item[\textbf{DQN~\cite{DQN}:}] Based on the DQN near-field dispatching algorithm, vehicles are matched only with requests from the dispatched region.

\item[\textbf{Random:}] Based on the dispatching framework we established, a randomized dispatching strategy is used here to compare the effectiveness of the decision algorithms.

\item[\textbf{NOD:}] The designed near-field dispatching framework is not used, and vehicles are matched only with the nearest requests (Further reference can be found in the paper~\cite{MARLODM}.).

\item[\textbf{NeurADP~\cite{shah2020neural}:}]  NeurADP calculates the expected future value of each vehicle assignment by using a neural network-based value function in an approximate dynamic algorithm.

\item[\textbf{DQN+MI:}] Based on DQN, the MI between vehicles and requests is considered as the intrinsic reward value of the DQN agents.

\item[\textbf{MFQL~\cite{pmlr-v80-yang18d}:}] The DQN algorithm is optimized by replacing the independently executed DQN algorithm with MFQL that considers the actions of surrounding vehicles to optimize the collaboration.

\item[\textbf{MFQL+MI:}] Based on MFQL, the MI between vehicles and requests is considered as the intrinsic reward value of MFQL agents.
\squishend

\begin{table}[htbp]
\centering
\setlength\tabcolsep{3pt}
\begin{tabular}{lllllllllll} 
    \hline
    {Varying}  & {DQN}  & {DQN+MI} & {MFQL}  &{MFQL+MI} \\
    \hline
    Request Num.                 &  6036.0 	       &  6221.6  	    &6077.8       &6219.2  \\
    Total revenue                                                                  	& 107571.97  &  110672.99  	&108190.70  &111326.63 \\
    \hline
    
    \hline
\end{tabular}
\caption{The effect of the mean field on the results under the same settings (NV=600, PD=300, C=4).}
\label{tab:MFQL_DQN}
\end{table}

\begin{table}[htbp]
\centering
\setlength\tabcolsep{3pt}
\begin{tabular}{lllllllllll} 
    \hline
    \multirow{2}{*}{Varying} & \multicolumn{3}{c}{Parameters} & \multirow{2}{*}{NeurADP~\cite{shah2020neural}}  & \multirow{2}{*}{MFQL+MI} &	\multirow{2}{*}{Improve} \\
    \cline{2-4}
                                     & NV    & PD  & C   \\
    \hline
    \multirow{3}{*}{PD}                 &  600	   &  100	 & 4	& 76061.72 	   &  77849.47  	&2.35\% \\
                                        &  600	   &  200	 & 4	& 96037.67 	   &  97959.34  	&2.00\% \\
                                        &  600	   &  300	 & 4	& 106180.23 	   &  110196.19  	&3.78\% \\
    \hline                                                                                                                 
    \multirow{3}{*}{NV}                 &  300	   &  300	 & 4	& 52125.31 	   &  53879.17  	&3.36\% \\
                                        &  400	   &  300	 & 4	& 72020.79	   &  74147.34  	&2.95\% \\
                                        &  500	   &  300	 & 4	& 89003.86 	   &  92292.44  	&3.69\% \\
    \hline                                                                                                                 
    \multirow{3}{*}{C}                  &  600	   &  300	 & 2	& 74844.92 	   &  76693.32  	&2.47\% \\
                                        &  600	   &  300	 & 4	& 107632.17 	   &  111326.63 	&3.43\% \\
                                        &  600	   &  300	 & 8	& 117115.35 	   &  120660.29  	&3.03\% \\
    \hline
    
    \hline
\end{tabular}
\caption{The total revenue. We compared MFQL+MI with NeurADP~\cite{shah2020neural}.}

\label{tab:revenue_impactNeurADP}
\end{table}

\subsection{Results}


The request data we used for training is from March 23 to April 1, 2016, 8 weekdays, and the validation data is used from April 22, 2016. The request data for testing is from April 4 to 8, 2016 for 5 weekdays. The location of the vehicle is initialized randomly at the beginning of individual instances in both training and testing phases. The decision duration $\Delta$ is set as 60s by default.

Since the algorithm is more necessary for the rational dispatching of vehicles during peak hours, we tested the trained model with the evening peak hours of 18-19 in Table~\ref{tab:pricing_impact_DQN}-\ref{tab:MFQL_DQN}. It is also repeated on multiple weekdays, and the results are averaged over 5 runs.

To illustrate the performance of our proposed framework, we compared them in different dimensions. One of the metrics is the revenue value, this value is the total revenue of all vehicles during the peak hours, which is proportional to distance and time according to Uber's charging strategy. The second metric is the total number of service requests, which is also the total number of requests served by the platform. 

In the end, we visualized the comparison of the data for the full 24-hour period, as shown in Figure~\ref{fig:DQN_oneday} and Figure~\ref{fig:MFQL_oneday}. We also visualized the distribution of peak hours the distribution of vehicles and requests as shown in 
Figure~\ref{fig:MutualInformation_MFQL}. 
The following are the conclusions of the key questions from the analysis of the results:
\squishlist
\item 1. (Dispatching framework effectiveness) From Table~\ref{tab:pricing_impact_DQN}, our proposed dispatching framework based on the simplest DQN algorithm has an improvement of 5\% even in the worst case compared to 
the nearest neighbor matching algorithm (NOD). At the same time, we can see that the best improvement can reach 28.9\%, and the improvement is greater as more vehicles and capacity are provided by our system, because the relatively large number of vehicles requires the algorithm to coordinate the dispatching on the map. In addition, Compared to the random dispatching algorithm (Random), the algorithm using DQN achieves higher results in all cases (Different number of vehicles and capacity). These results, all illustrate the effectiveness of our proposed RL-based vehicle dispatching framework.

\item 2. (The advantage of the intrinsic reward of MI) With other components unchanged, in Table~\ref{tab:pricing_impact_DQN}, comparing the two algorithms DQN+MI (DQN algorithm considering MI between vehicle distribution and order distribution) and DQN (DQN algorithm without considering MI), the overall revenue DQN+MI can improve up to 2.88\% relative to the DQN algorithm. In Table~\ref{tab:pricing_impactMFQL}, comparing MFQL+MI (MFQL algorithm considering MI) with MFQL (MFQL algorithm without considering MI) the overall revenue can be improved by up to 3.53\%. In a market of hundreds of millions of dollars, a 1\% improvement is already significant~\cite{Xu2018}. These results illustrate that the reward value based on MI between vehicles and orders further optimizes the RL-based dispatching strategy. Also as shown in Appendix Figure \ref{fig:DQN_oneday} and Appendix Figure \ref{fig:MFQL_oneday}, we visualize the revenue graph for a run in a day, and these results show that the MI module increases the overall revenue.

\item 3. (The advantage of Mean Field) With the other components unchanged, allowing the algorithm to consider the effect of neighboring actions sets the MFQL algorithm. In Table~\ref{tab:MFQL_DQN}, by comparing MFQL with DQN, and MFQL+MI with DQN+MI we can see that the algorithm with MFQL has about 0.6\% improvement.

\item 4. Compared with the existing best method for on-demand ride pooling NeurADP~\cite{shah2020neural} in Table~\ref{tab:revenue_impactNeurADP}, our algorithm has an average performance improvement of 3\%.

\item 5. We visualize the curve of the value of the MI during the training process, as shown in Figure~\ref{fig:MutualInformation_value} It can be seen that the MI gradually increases and stabilizes as the training progresses. This indicates that the distribution of the vehicles becomes more dependent on the distribution of the requests as the training progresses.

\item 6. We visualize the distribution of vehicles (cumulative distribution over the time of peak hours) in the case of using MFQL algorithm and MFQL+MI algorithm as shown in Appendix Figure~\ref{fig:MutualInformation_MFQL}. The distribution of the pickup locations of the request is shown in Figure~\ref{fig:MutualInformation_MFQL}(a), and the distribution of the drop-off locations of the request is shown in Figure~\ref{fig:MutualInformation_MFQL}(b). The difference distribution of vehicles due to the two algorithms is shown in  Figure~\ref{fig:MutualInformation_MFQL}(d).
We can see from the difference distribution that the MFQL+MI algorithm makes more vehicles to be distributed at the locations where the requests appear. 
Similarly, we visualize the distribution of vehicles at peak hours 18-19 with the DQN and DQN+MI algorithms, as shown in  Figure~\ref{fig:MutualInformation_MFQL}(c). We can see that the distribution of vehicles is closer to the distribution of requests for the algorithm using the MI component. Combined with the revenue results we obtained previously, we can know that our MI component makes the distribution of vehicles more reasonable and thus enables the revenue higher.

\item 7. (The effectiveness of the pickup delay time) We tested our algorithm in different environmental conditions with different pickup delay times. As can be seen in Table~\ref{tab:pricing_impact_DQN} and Table~\ref{tab:pricing_impactMFQL}, our algorithm MFQL+MI achieved better results in both cases.

\item 8. (The effectiveness of the capacity) Another factor is the capacity, where MFQL+MI is significantly dominated by the other methods in terms of both revenue value and the number of requests for different capacity cases, as can be seen in Table~\ref{tab:pricing_impact_DQN} and Table~\ref{tab:pricing_impactMFQL}.

\item 9. To setup the parameters for the number of neighbors in the MFQL algorithm to obtain the best results, we tested different numbers of neighbors. As shown in Table~\ref{tab:differentNumberNeighbor}, the algorithm achieves the best results when the number is 6.

\item 10. For setting the parameters of MI to obtain the best results, we tested the effectiveness of the algorithm for different $\ddot{\alpha}$ cases. As shown in Table~\ref{tab:differentcoeffi}, the algorithm achieves the best results when $\ddot{\alpha}$ is 0.09.
\squishend

\section{Conclusion}
Companies such as UberPool and LyftLine offer ride pooling services that allow people to travel at a lower cost, while also increasing driver revenue. Most existing works focus on improving overall revenue (overall platform revenue or driver revenue), ignoring consideration of order distribution and vehicle distribution, often resulting in unusually distributed requests that are difficult to get a ride. To make the algorithm applicable to on-demand rides, and to consider future benefits and vehicle-request distribution differences in dispatching and matching decisions, we propose a RL-based vehicle dispatching framework. The algorithm uses the mutual information (MI) between vehicle and request distribution as the intrinsic reward value for vehicle dispatching. Our approach consistently outperforms the existing baseline, and in the best case, improves revenue by 3\% for a city-scale taxi dataset over existing best method for on-demand ride pooling. Considering that even a 1\% improvement in revenue is considered by the industry to be a large improvement on similar transportation problems ~\cite{Xu2018,Lowalekar2019ZACAZ}.
\newpage

\balance
\clearpage
\bibliographystyle{ACM-Reference-Format}
\bibliography{sample-base}

\newpage
\appendix

\section{Algorithm}

\begin{algorithm}[htbp]
\caption{Dispatching and matching} 
\label{alg:algorithm1}
\begin{algorithmic}[1] 
\STATE Initialize: replay buffer \emph{M}, $Q_v$ function.
\FOR {each training episode  $0 \leq \emph{epi} < N$ }
\STATE Reset environment and vehicle location, and get intial $\omega_v$ for each vehicle $v$
\FOR {$t = 0 \dots$ steps per \emph{epi}}
\STATE Choose actions $a_v  \sim \pi_{v}( \cdot \mid \omega_{v},{\bar{a}}_{N_v})$ for each vehicle (Equation~\eqref{eqn:actionsel})
\STATE The obtained action $a_v$ is the area where the vehicle is dispatched
\STATE According to the restricted region $a_v$, the system obtains the set of available request combinations $\mathcal{F}_v^t$ for the vehicle $v$ (Equation~\eqref{eqn:reqcomb})
\STATE Matching using integer programming linear algorithm (Equation~\eqref{eqn:ILPobj})
\STATE Get reward $r_v$ and next state ${\omega\prime}_v$ for all vehicles
\STATE Computing $I(V_D;E_D )$ as an intrinsic reward, obtaining the total reward $r_v + I(V_D;E_D)$
\STATE Add episode to buffer, \emph{M}
\IF {$ t \geq$ BatchSize}
\STATE Sample minibatch, $B$
\STATE Q function update according to Equation~\eqref{eqn:mfqlup}
\ENDIF
\ENDFOR
\ENDFOR
\end{algorithmic}
\end{algorithm}

\section{Mutual Information}

\begin{equation}
\begin{aligned}
I\left(V_D ; E_D\right) & =\iint p\left(v_d, e_d\right) \log \frac{p\left(v_d, e_d\right)}{p\left(v_d\right) p\left(e_d\right)} \\
& = \iint p\left(v_d, e_d\right) \log \frac{p\left(v_d \mid e_d\right)}{p\left(v_d\right)} \\
& =\iint p\left(v_d, e_d\right) \log \frac{q\left(v_d \mid e_d\right)}{p\left(v_d\right)} \\
& +D_{k l}\left(p\left(v_d \mid e_d\right) \| q\left(v_d \mid e_d\right)\right) \\
& \geq \iint p\left(v_d, e_d\right) \log \frac{q\left(v_d \mid e_d\right)}{p\left(v_d\right)} \\
& \geq \iint p\left(v_d, e_d\right) \log \frac{q\left(v_d \mid e_d\right)}{p\left(v_d\right)} \\
& =\! \! \! \iint \! \! p\left(v_d, e_d\right) \! \log \! q\left(v_d \! \mid \! e_d\right) \! - \!\! \int \! p\left(v_d\right) \! \log \! p \!\left(v_d\right) \\
& =\iint p\left(v_d, e_d\right) \log q\left(v_d \mid e_d\right)-\int p\left(v_d\right) \log p\left(v_d\right) \\
& = \! \int \! p\left(e_d\right) \int p\left(v_d \mid e_d\right) \! \log \! q\left( \! v_d  \! \mid \! e_d\right) \! + \! H \! \left(v_d\right) \\
& =\!\! -\mathbb{E}_{e_d \sim \! E_D} \! \left[CE\!\left[p\left(v_d \! \mid  \! e_d\right) \! \| q\left(v_d \mid e_d\right)\right] \!\! + \!\! H \! \left(v_d\right)\right.
\end{aligned}
\label{eqa:MI}
\end{equation}

\section{Related Work}

How to increase the revenue of drivers while balancing the supply and demand relationship between passengers and vehicles, and improving the user experience of drivers and passengers is an important content of taxi platforms \cite{Large-ScaleFleet18}. Traditionally, researchers study the supply-demand balance of drivers services \cite{Spatio-Temporal9}. The development of mobile internet technology makes it easier to collect driver data, which leads to the emergence of many data-driven methods \cite{Data-Driven24,Cost-Effective31,T-Finder49,chen2022curiosity}. By using simple data methods, we can directly analyze the patterns of historical driver data and drivers can search for orders along a route rich in passengers \cite{Cost-Effective31}. It is also possible to identify hotspots for drivers to stay in these areas \cite{T-Finder49}. However, these methods lack coordination between drivers. In addition, there are also some methods that use models to directly model drivers and passengers, simulating the supply and demand relationship between drivers and orders. These methods can be solved using combinatorial optimization \cite{PrivateHunt42} or mixed integer programming \cite{Dispatch44} to dispatch vehicles. These methods rely on established models and cannot adapt well to dynamic environments \cite{MOVI30}.

Recently, researchers utilize reinforcement learning algorithms to alleviate the supply-demand relationship between taxis and orders \cite{Spatio-Temporal9, Large-ScaleFleet18, Collaborative33, MOVI30}. These algorithms do not rely on models and can directly learn dispatching strategies in dynamic environments. To integrate complex order and vehicle information in cities, these methods typically utilize deep learning methods to accelerate strategy learning. Xu et al. \cite{Xu2018} and Tang et al. \cite{value-Network2} used policy evaluation methods to derive region-specific time-varying value functions to assist in matching between orders and vehicles. They modeled vehicle focus points as Markov decision processes (MDPs) and utilized various reinforcement learning algorithms, such as the DQN algorithm based on learning the value function in the supply and demand heat map \cite{MOVI30}, the Deep Q-Network (DQN) considering contextual features \cite{NEURIPS2018_94bb077f10}, and hierarchical reinforcement learning for order dispatching and fleet management \cite{CoRide}.
Although these works do improve system performance in dynamic environments compared to traditional methods, the overall performance improvement is limited due to the simple design of reinforcement learning reward values, which can not capture the supply-demand relationship between orders and vehicles. In this paper, our proposed algorithm takes the mutual information between request and vehicle distribution as the reward value for reinforcement learning agents, further improving the revenue of taxi platforms.

\newpage
\onecolumn

\section{Compare the differences in the distribution and results}

\begin{figure*}[ht]
    \centering
    \includegraphics[width=0.9\linewidth]{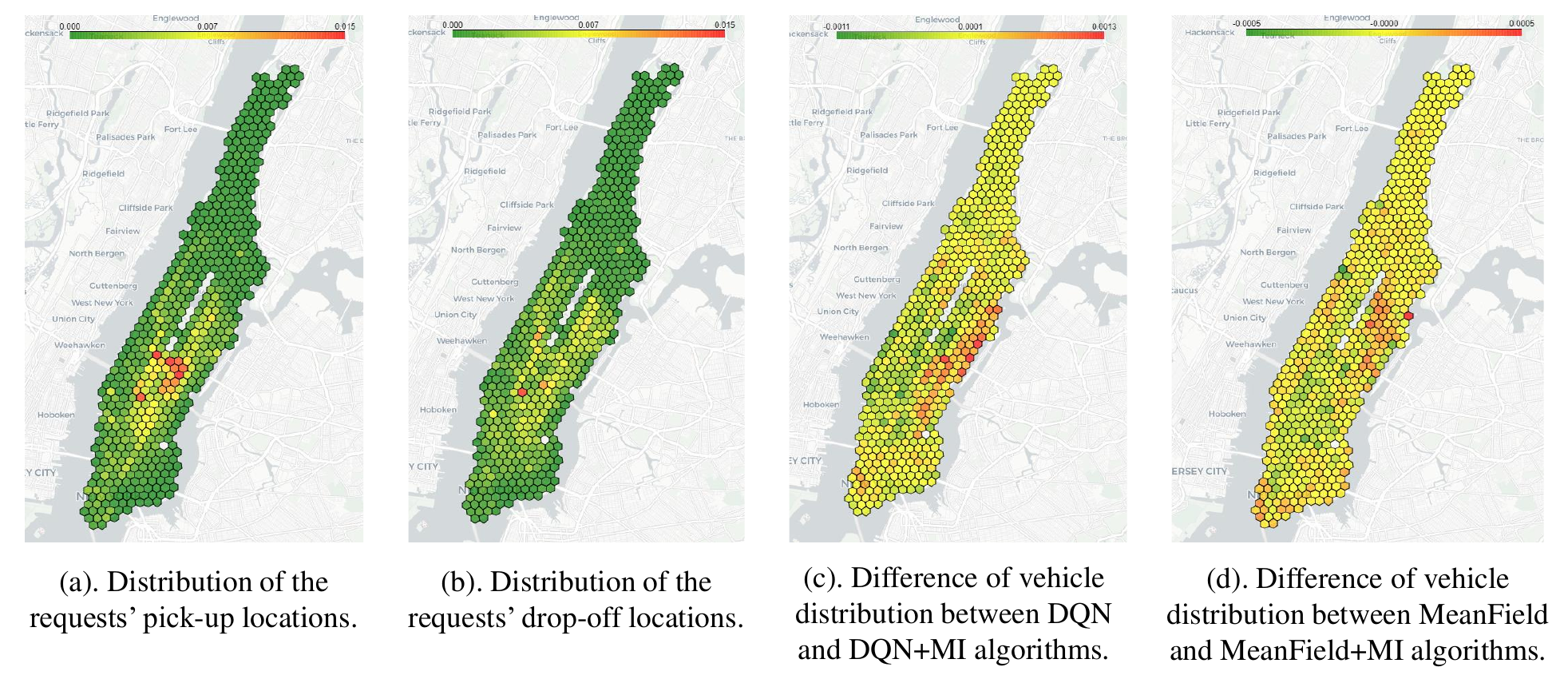}
    \caption{Compare the differences in the distribution of vehicles with and without MI modules for different algorithms. (a) and (b) are the distribution of requests. (c) is the difference in the distribution of vehicles, which is calculated by computing the distribution of vehicles under the DQN+MI algorithm minus the distribution under the DQN algorithm. (d) is under MFQL+MI minus MFQL algorithm.}
    \label{fig:MutualInformation_MFQL}
\end{figure*}

\begin{figure*}[htbp]
\centering    
\subfigure[600 vehicles, capacity of 4.] 
{
	\begin{minipage}{0.32\linewidth}
	\centering          
	\includegraphics[width=2.5in,height=2in]{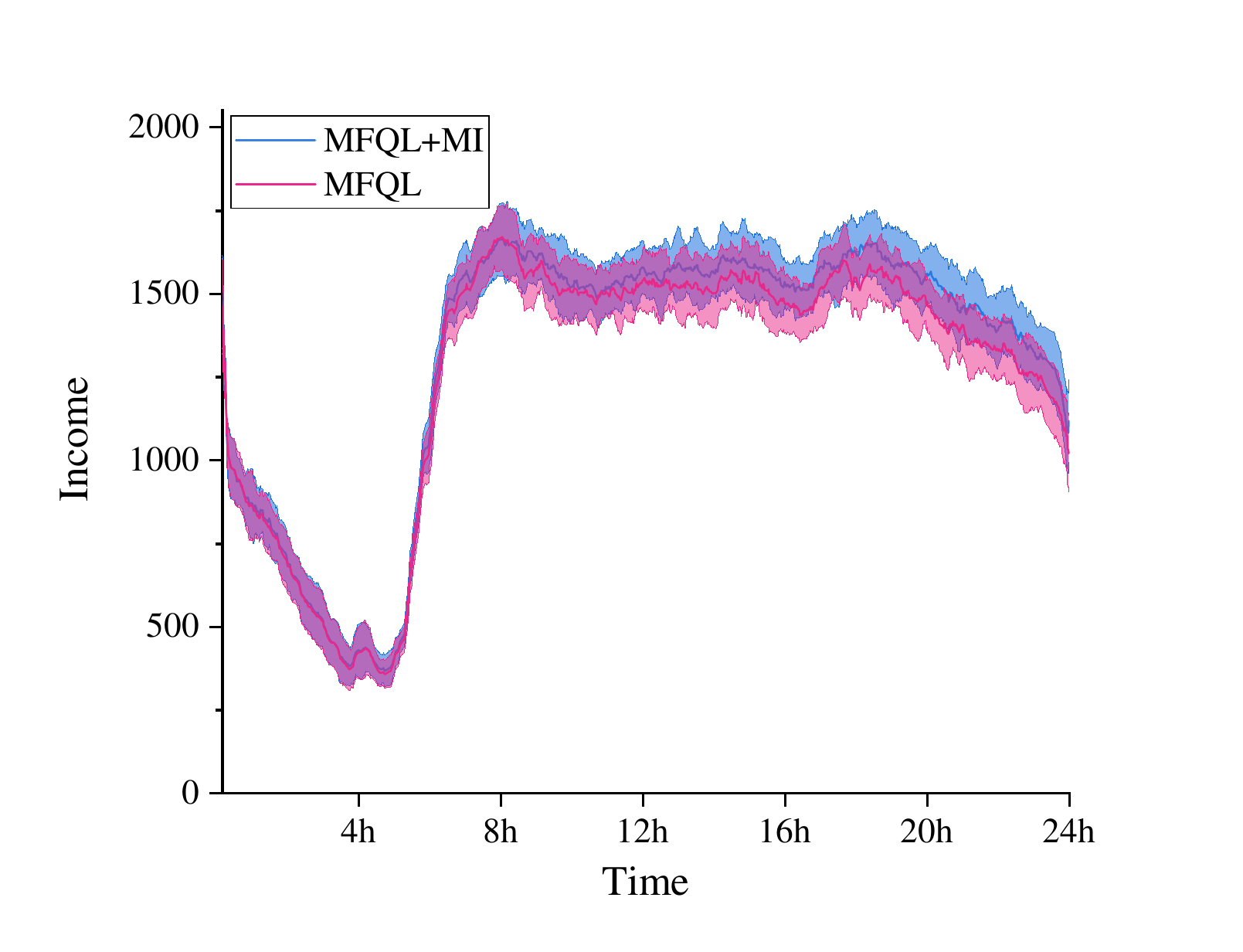}   
	\end{minipage}
}
\subfigure[500 vehicles, capacity of 4.] 
{
	\begin{minipage}{0.32\linewidth}
	\centering      
	\includegraphics[width=2.5in,height=2in]{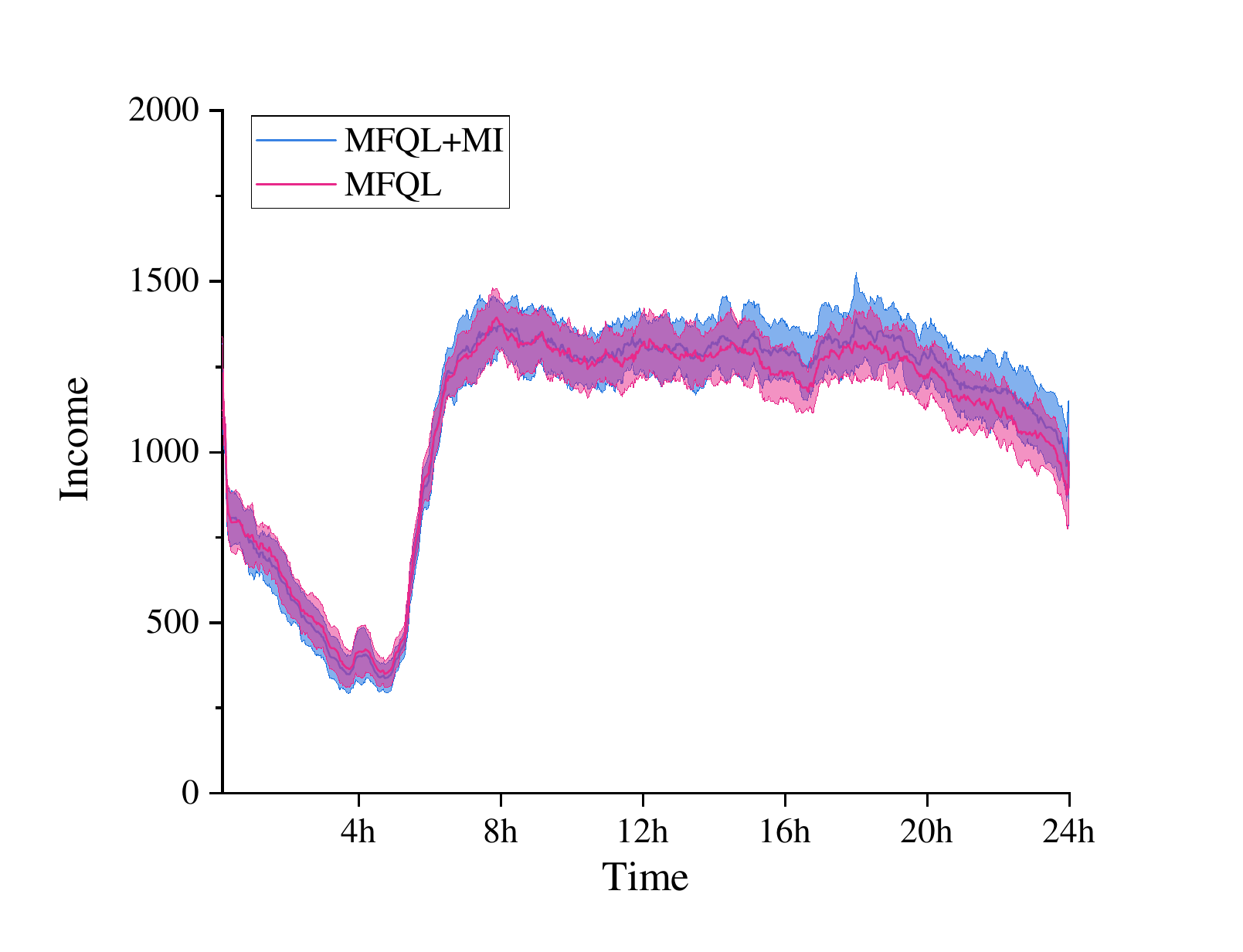}   
	\end{minipage}
}
\subfigure[400 vehicles, capacity of 4.] 
{
	\begin{minipage}{0.32\linewidth}
	\centering      
	\includegraphics[width=2.5in,height=2in]{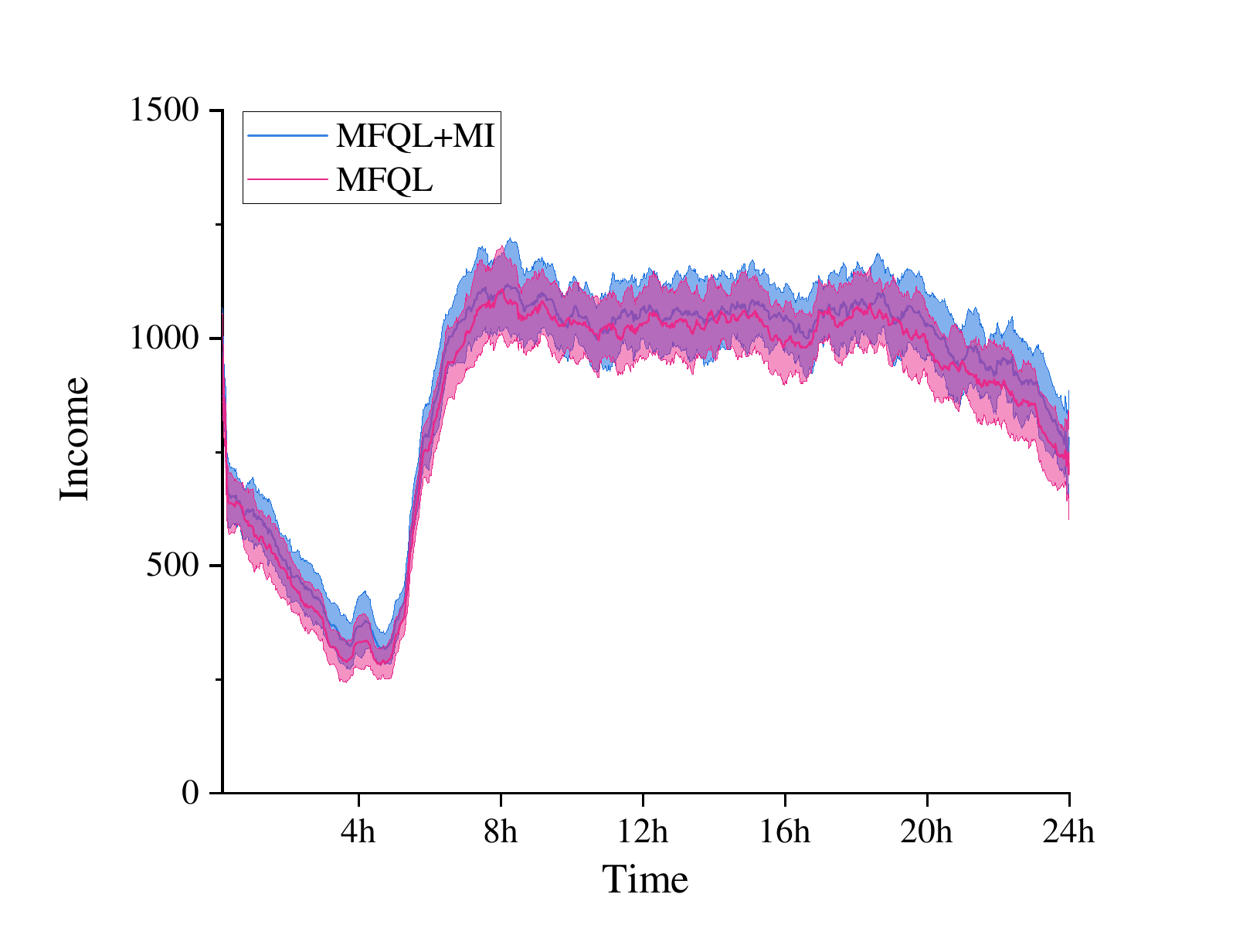}   
	\end{minipage}
}
\caption{The MI module is added to the MFQL and the revenue curve of the results of the whole day running.} 
\label{fig:MFQL_oneday}  
\end{figure*}

\end{document}